\documentclass[sn-mathphys-num]{sn-jnl}% 
\usepackage{graphicx}%
\usepackage{multirow}%
\usepackage{amsmath,amssymb,amsfonts}%
\usepackage{amsthm}%
\usepackage{mathrsfs}%
\usepackage[title]{appendix}%
\usepackage{xcolor}%
\usepackage{textcomp}%
\usepackage{manyfoot}%
\usepackage{booktabs}%
\usepackage{algorithm}%
\usepackage{algorithmicx}%
\usepackage{algpseudocode}%
\usepackage{listings}%
\theoremstyle{thmstyleone}%
\theoremstyle{thmstyletwo}%

\theoremstyle{thmstylethree}%

\begin{document}

\title[Article Title]{A-MFST: Adaptive Multi-Flow Sparse Tracker for Real-Time Tissue Tracking Under Occlusion}

%%=============================================================%%
%% GivenName	-> \fnm{Joergen W.}
%% Particle	-> \spfx{van der} -> surname prefix
%% FamilyName	-> \sur{Ploeg}
%% Suffix	-> \sfx{IV}
%% \author*[1,2]{\fnm{Joergen W.} \spfx{van der} \sur{Ploeg} 
%%  \sfx{IV}}\email{iauthor@gmail.com}
%%=============================================================%%

\author*[1]{\fnm{Yuxin} \sur{Chen}}\email{yuxinchen@ece.ubc.ca}

\author[1]{\fnm{Zijian} \sur{Wu}}\email{zijianwu@ece.ubc.ca}
% \equalcont{These authors contributed equally to this work.}

\author[2]{\fnm{Adam} \sur{Schmidt}}\email{Adam.Schmidt@intusurg.com}
% \equalcont{These authors contributed equally to this work.}

\author[1]{\fnm{Septimiu E.} \sur{Salcudean}}\email{tims@ece.ubc.ca}

\affil*[1]{\orgdiv{Department of Electrical and Computer Engineering}, \orgname{The University of British
Columbia}, \orgaddress{\city{Vancouver}, \postcode{V6T 1Z4}, \state{BC}, \country{Canada}}}

\affil[2]{\orgname{Intuitive Surgical Inc.}, \orgaddress{\city{Sunnyvale}, \postcode{94086}, \state{CA}, \country{United States}}}

% \affil[3]{\orgdiv{Department}, \orgname{Organization}, \orgaddress{\street{Street}, \city{City}, \postcode{610101}, \state{State}, \country{Country}}}

%%==================================%%
%% Sample for unstructured abstract %%
%%==================================%%

\abstract{
\textbf{\\* Purpose}:
Tissue tracking is critical for downstream tasks in robot-assisted surgery. The Sparse Efficient Neural Depth and Deformation (SENDD) model has previously demonstrated accurate and real-time sparse point tracking, but struggled with occlusion handling. This work extends SENDD to enhance occlusion detection and tracking consistency while maintaining real-time performance.
%\smallskip
\textbf{\\* Methods}: 
We use the Segment Anything Model2~(SAM2)~\cite{ravi2024sam} to detect and mask occlusions by surgical tools,
and we develop and integrate into SENDD an Adaptive Multi-Flow Sparse Tracker (A-MFST) with forward-backward consistency metrics, to enhance occlusion and uncertainty estimation. 
A-MFST is an unsupervised variant of the Multi-Flow dense Tracker (MFT)~\cite{neoral2024mft}.
%\smallskip
\textbf{\\* Results}: 
We evaluate our approach on the STIR dataset~\cite{schmidt2023stir}, and demonstrate a significant improvement in tracking accuracy under occlusion, reducing average tracking errors by 12\% in Mean Endpoint Error~(MEE) and showing a 6\% improvement in \( \delta_{\text{avg}}^{x} \), the averaged accuracy over thresholds of [4, 8, 16, 32, 64] pixels~\cite{doersch2022tap}. The incorporation of forward-backward consistency further improves the selection of optimal tracking paths, reducing drift and enhancing robustness. Notably, these improvements were achieved without compromising the model’s real-time capabilities.
%\smallskip
\textbf{\\* Conclusions}: 
Using A-MFST and SAM2, we enhance SENDD's ability to track tissue in real-time, under instrument and tissue occlusions.}

%%\pacs[JEL Classification]{D8, H51}

%%\pacs[MSC Classification]{35A01, 65L10, 65L12, 65L20, 65L70}

\maketitle

\section{Introduction}\label{sec1}

Tissue tracking has many applications in robotic-assisted surgery (RAS)~\cite{schmidt2024tracking}, e.g., maintaining tissue registration to pre-operative imaging for augmented reality. 
%potentially leading to misalignment or loss of crucial information. In RAS, such inaccuracies can increase the challenge of applying accurate downstream applications, emphasizing the need for robust, real-time tissue tracking systems capable of handling occlusions effectively.
Since both real-time performance and high accuracy are important, 
%Tracking systems must update tissue and tool positions with minimal latency to provide surgeons with real-time visual feedback. Achieving this balance is complicated by the inherent trade-off between computational efficiency and precision. 
%Speed is crucial to avoid delays in visualization, while accuracy ensures that key anatomical features are faithfully tracked throughout the procedure. 
sparse point tracking has been identified as a promising tissue tracking methods, as it reduces the number of points that need to be processed.  
%thereby increasing speed without compromising essential tracking information. 
%Sparse points focus on key features, which are vital for registration and alignment during surgery.
The Sparse Efficient Neural Depth and Deformation (SENDD)~\cite{schmidt2023sendd} model 
%is one of the state-of-the-art methods that addresses the need for real-time performance in RAS. SENDD
is a highly efficient and accurate solution for 3D tracking of tissue keypoints in stereo endoscopy, 
%of real-time applications, 
making it an attractive option for clinical deployment. 
%By tracking sparse points, SENDD minimizes the computational load while focusing on key anatomical structures that are essential for maintaining accurate tissue registration throughout the surgery.
Significant challenges in tissue tracking are tracking drift and the presence of occlusions, caused by surgical instruments or by tissue folding onto itself~\cite{schmidt2024tracking}.  
%Occlusions hinder a tracking system's ability to maintain consistent and reliable performance. 
Like other tracking methods, SENDD does not address these challenges. Tracking drift occurs when small inaccuracies accumulate over time, causing the tracked points to deviate from their true positions. When surgical instruments block the view of the tissue, the model can generate erroneous updates and lose key points.

\begin{figure}[tbh]
\centering
\includegraphics[width=0.9\textwidth]{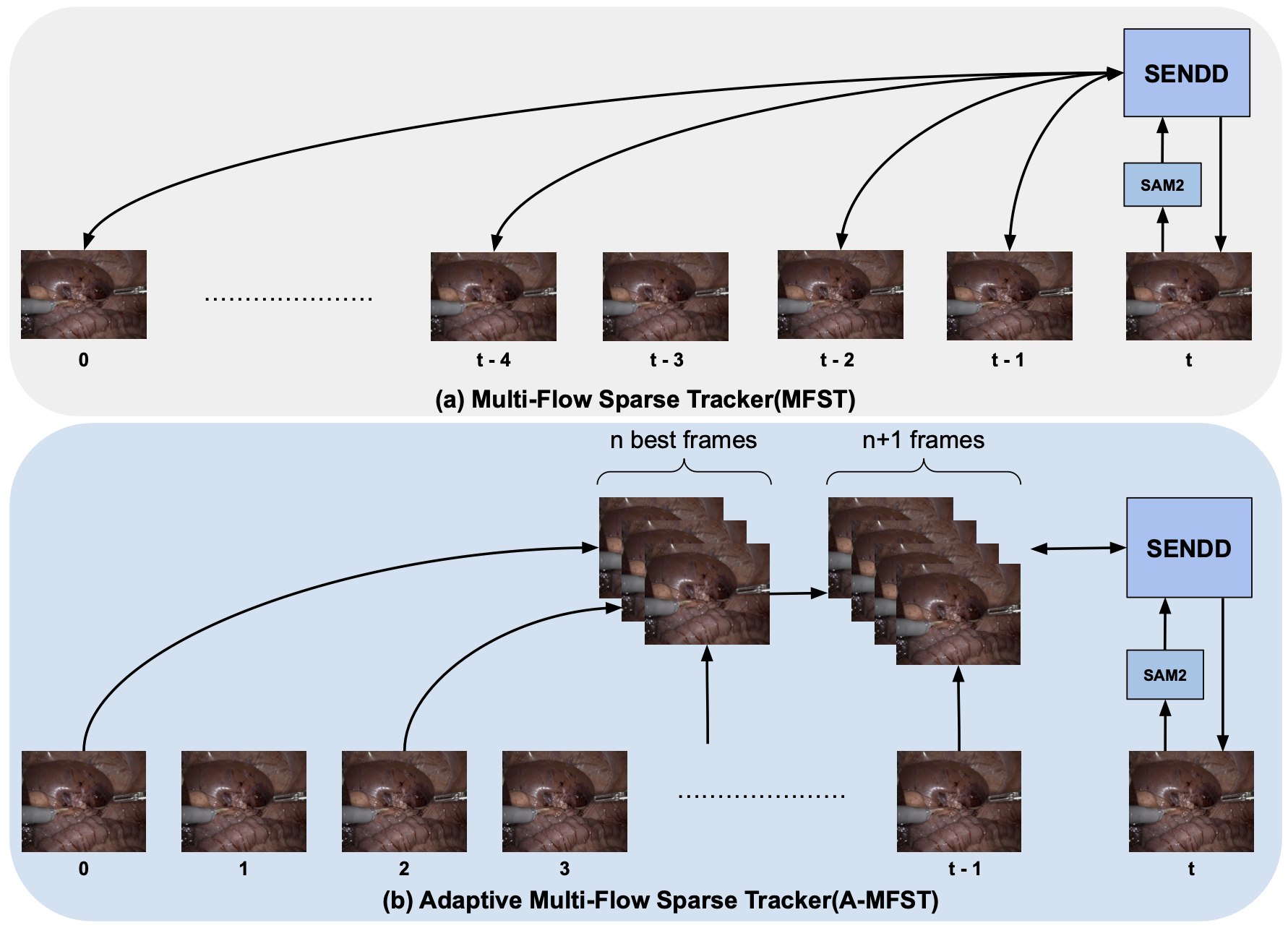}
\caption{Overall structure of the tracking algorithm: (a) Multi-Flow Sparse Tracker (MFST) (b) Adaptive Multi-Flow Sparse Tracker (A-MFST)}\label{overall_structure}
\end{figure}

This paper proposes an enhanced version of the SENDD model, with the following contributions:
%that improves its ability to handle occlusions while retaining the ability for real-time tracking. We introduce several key modifications aimed at addressing the challenges posed by occlusions and drift. 
First, we incorporate a state-of-the-art segmentation model, SAM2~\cite{ravi2024sam}, to segment surgical instruments,
%SAM2 enables the detection of occlusions as they occur, 
preventing the model from making erroneous updates when tissue is blocked by instruments from view. 
Second, we develop Multi-Flow Sparse Tracker (MFST) a training-free variant of the Multi-Flow dense Tracker (MFT)~\cite{neoral2024mft} framework to improve long-term tracking performance. 
Third, we propose A-MFST, an adaptive frame selection extension of MFST (Fig.~\ref{overall_structure}). 
While the original MFT relies on CNNs trained on the Kubric dataset~\cite{greff2021kubric}, which includes ground truth occlusion labels for each point in every frame, A-MFST can dynamically select the optimal frames for back-checking without requiring training or ground truth labels—both of which are limited in endoscopic environments. A-MFST maintains robust tracking through medium-length occlusions, reducing drift and enhancing tracking accuracy over extended periods.
These enhancements to the SENDD model create a more resilient tissue-tracking framework that addresses the critical challenges posed by occlusions and drift while retaining real-time tracking ability. 
\section{Related Work}\label{sec2}
%
%\subsection{Tissue Tracking in Robotic Surgery}\label{subsec2-1}
%Tracking deformable,  tissues presents significant challenges due to their complex motion and shape changes during surgery. 
Early approaches \cite{grasa2011ekf, grasa2013visual} to tissue tracking assumed rigid tissue motion. More recent methods incorporate deformable models and simultaneous localization and mapping algorithms~\cite{song2018mis}. As outlined in \cite{schmidt2024tracking}, deformable tissue tracking methods  include optical flow, feature matching, and machine learning-based models. 
The Sparse Efficient Neural Depth and Deformation (SENDD)~\cite{schmidt2023sendd} model is self-supervised and it achieves accurate, real-time, 3D tissue tracking by focusing on key anatomical landmarks.
%
%\subsection{Occlusion Handling Strategies}\label{subsec2-2}
%
Traditional methods for occlusion handling in optical flow typically rely on heuristics such as temporal smoothness or motion segmentation. For example, methods such as TV-L1 optical flow~\cite{zhang2017robust} identify discontinuities in motion, where sudden changes in pixel flow are interpreted as occlusions. 
These methods can handle simple occlusions, but they often fail for complex or long-term occlusions.
In recent work, instance, RAFT \cite{teed2020raft} and FlowFormer \cite{huang2022flowformer} estimate dense optical flow by constructing correlation volumes between image pairs. Occlusions are not specifically detected and are managed by refining optical flow estimates through iterative updates. 
Ada-Tracker~\cite{guo2024ada} employs an adaptive template updating mechanism that refines inter-frame optical flow estimates, using confidence metrics to counteract occlusions and drift. 
PIPs++~\cite{zheng2023pointodyssey} and PIPsUS~\cite{chen2024pipsus} tackles occlusions by extending the temporal receptive field for point tracking, allowing point tracks to update and recover over long video sequences. 
CoTracker~\cite{karaev2023cotracker} takes a joint approach by tracking spatially correlated points, enabling robust recovery from occlusions through collective tracking, but it can still cause drift. SpatialTracker~\cite{xiao2024spatialtracker} shifts 2D points into 3D space, enforcing rigid-body constraints to manage occlusions and complex motions. Finally, MFT~\cite{neoral2024mft} combines optical flow estimates from both consecutive and distant frames, selecting reliable paths to ensure long-term tracking and recovery from occlusions. 
%These approaches demonstrate the diverse ways to address occlusion challenges in tracking applications.
While these methods are effective in short-term occlusions, they struggle to recover from long-term occlusions and are prone to drift over time.
\section{Methods}
%
%This section presents SENDD with SAM2 and Adaptive Multi-Frame Sparse Tracker (A-MFST) framework. \\
%
\textbf{SAM2-Based Instrument Segmentation for Occlusion Detection:}
To effectively manage occlusions from instruments during tissue tracking, we use the Segment Anything Model2 (SAM2)~\cite{ravi2024sam}. 
%SAM2 can be used to generate a mask of surgical instruments in the video frames, enabling precise instrument occlusion detection. This segmented instrument mask informs the tracking algorithm which points are occluded by the instrument, ensuring that only visible points are updated during the tracking process.
%
\begin{figure}[t]
\centering
\includegraphics[width=0.9\textwidth]{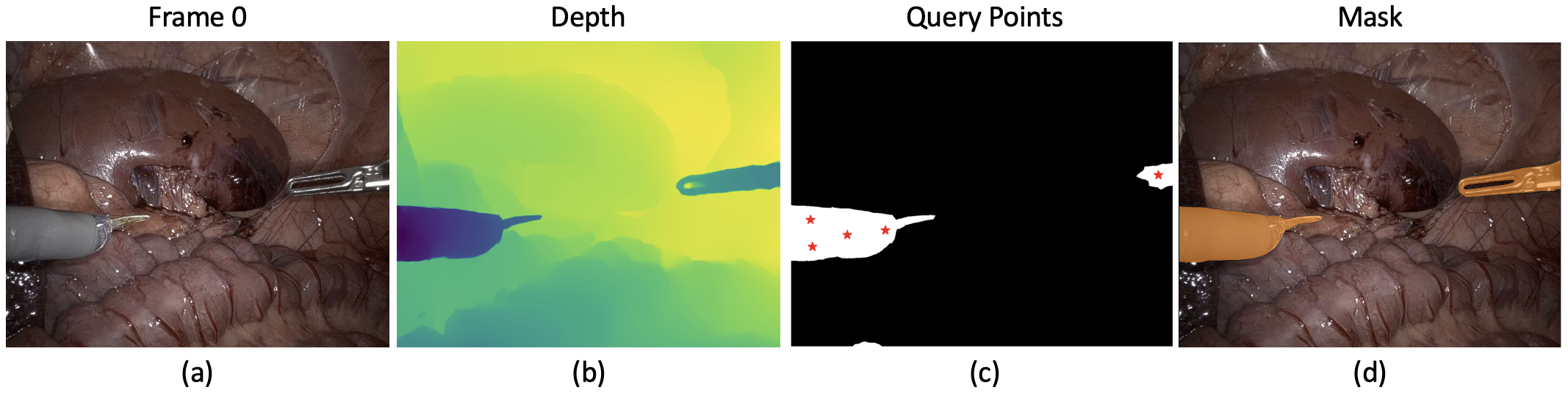}
\caption{Illustration of the Initialization Process for SAM2: (a) Original Image of Frame 0; (b) Depth Map; (c) Thresholded Depth and Initilized Query Points; and (d) Mask Labeled Image.}\label{SAM2-fig}
\end{figure}
%
%
%\subsubsection{Depth Estimation and Query Point Initialization for SAM2}
%
SAM2 requires initialization in the first frame by identifying key points on the instrument to generate an initial segmentation mask. To automate this process and reduce manual input, we adopt a depth-based method for selecting query points (Fig.~\ref{SAM2-fig}). Datasets such as the STIR dataset~\cite{schmidt2023stir} provide camera calibration parameters for stereo rectification and depth estimation.

Following rectification, we use RAFT~\cite{teed2020raft}, an optical flow model, to estimate the pixel disparity between the left and right stereo images and estimate depth. 
%RAFT computes the flow by matching pixel correspondences between the two frames, and the disparity between these matched points is then used to calculate depth.
%
Once the depth is computed for each pixel in the first frame, we apply a depth threshold to isolate the instrument from the surrounding tissue. 
%This threshold is set to exclude points that lie beyond a certain depth, ensuring that only the closer objects, typically the instruments, are selected. 
Inspired by~\cite{wu2024real}, the points within this depth range are then generated by K-Medoids clustering~\cite{mannor2011k} centers as query points to initialize SAM2 and generate the instrument mask for the first frame. Due to variations in lighting conditions and instrument positioning, manual adjustments may occasionally be necessary for optimal query point selection. 

%
%\subsubsection{Occlusion Detection via Segmentation Mask}
%
Once the instrument mask is initialized in the first frame and SAM2 is set up, the model continues to generate segmentation masks for subsequent frames.  All tracking points that fall within the segmented instrument mask are flagged as occluded. The tracking algorithm excludes these points from updates until they are no longer within the instrument mask, ensuring that no erroneous updates are made when points are hidden by the instrument.
%
%\subsection{Multi-Flow Sparse Tracker for Occlusion Detection and Uncertainty Estimation}\label{subsec3-1}

\textbf{Multi-Flow Sparse Tracker (MFST):}  In the original implementation of MFT~\cite{neoral2024mft}, the algorithm provided a robust mechanism for long-term tracking by evaluating multiple flow chains across logarithmically spaced intervals (1, 2, 4, 8, 16, 32, $\infty$). A convolutional neural network (CNN) was employed to estimate occlusion and uncertainty scores. By comparing the scores associated with each candidate flow, the most reliable flow path was selected, managing partial and temporary occlusions. 
%This selection process ensures that the algorithm chooses the most stable and accurate flow path for each key point, even in the presence of partial or temporary occlusions. 
\begin{figure}[tbh]
\centering
\includegraphics[width=0.9\textwidth]{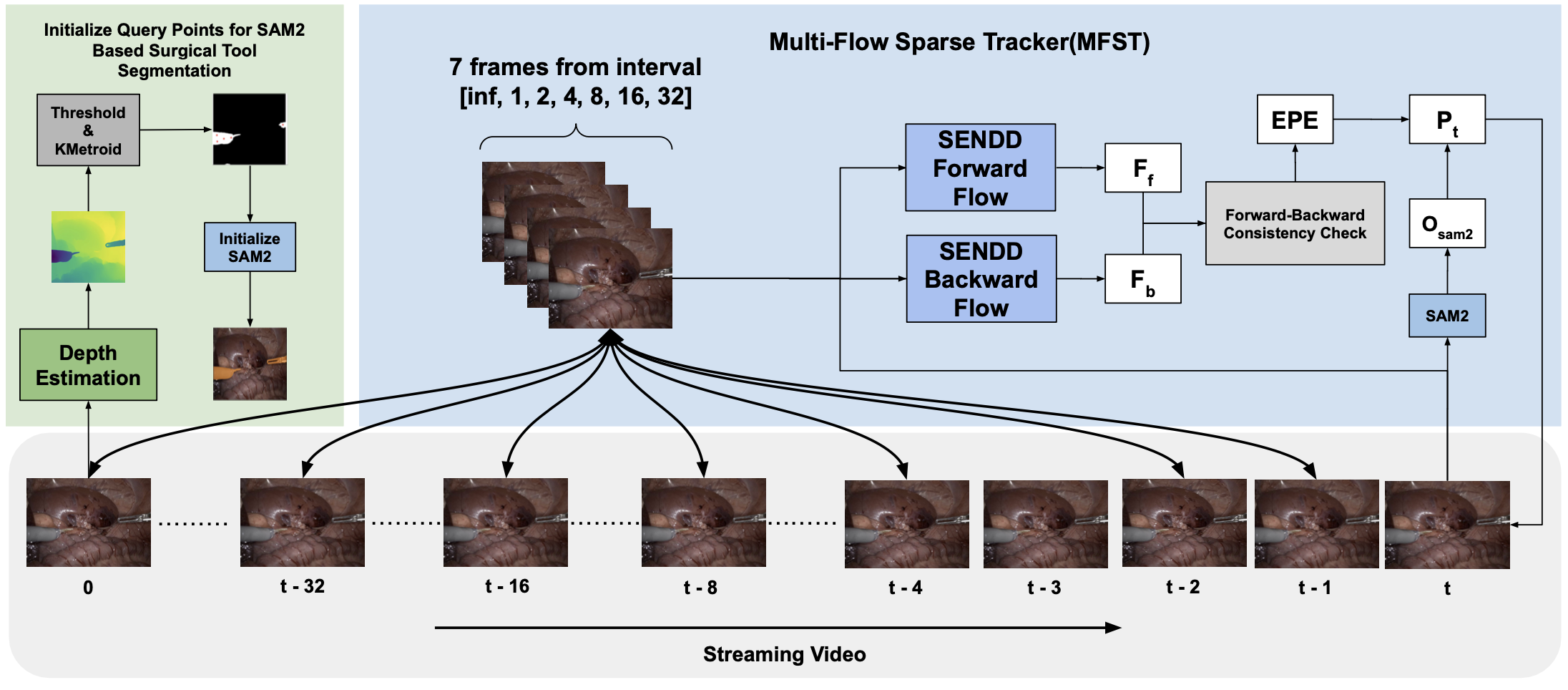}
\caption{Overall structure of the Multi-Flow Sparse Tracker (MFST)}\label{MFST_structure}
\end{figure}

To implement an MFT-like structure for sparse tracking, we propose the Multi-Flow Sparse Tracker (MFST) as shown in Figure~\ref{MFST_structure}. We replace the CNN with forward and backward consistency as metric to select the most reliable flow, which, unlike MFT, does not require ground truth for training.
%The occlusion and uncertainty CNN in MFT is trained using simulating ground truth, but we often do not have ground truth for training in endoscopy. 
%Thus, we choose to implement forward and backward consistency for selecting the best path from the multiple candidate flows.
%
%\subsubsection{Forward-Backward Consistency for General Occlusion Detection and Optimal Flow Path Selection}\label{subsubsec2}

%{\bf MSFT Forward-Backward Consistency:}
%Here, we apply forward-backward consistency checks in the MFST algorithm as metrics to select the best flow path. 
Forward-backward consistency compares the flow of points from SENDD between frames in both directions to assess tracking accuracy. While it is used to select the optimal path, it also works as another occlusion handling mechanism to address other forms of non-instrument occlusion, such as tissue overlapping.
To implement this, we calculate optical flow from SENDD between back-checked frames and the current frame. We then evaluate the consistency of the forward and backward flows by calculating the endpoint error (EPE) for each tracked point. If the EPE exceeds an empirically set threshold $\tau$, the point is considered occluded; the lowest EPE candidate is picked for the optimal path. This threshold $\tau$ was fine-tuned based on experimental results and is applied consistently across all sequences.
% We empirically set an endpoint error threshold $\tau$ to detect occluded points.
Unlike MFT, which saves the image for each frame and reuses it during back-checks, MFST stores the multi-scale global features computed from each frame. This eliminates the need to recalculate these features during the forward and backward consistency checks, improving efficiency by avoiding redundant computations.

By combining SAM2 segmentation-based occlusion detection with the forward-backward consistency checks provided by MFST, we create a comprehensive framework for occlusion handling. SAM2 effectively identifies instrument occlusions, while forward and backward consistency helps to select the optimal path and detects other occlusions, such as those caused by overlapping tissue or unpredictable movements. 
%This approach allows for more accurate tracking updates, minimizing drift and ensuring that the tracked points are updated only when visible. This integration significantly enhances SENDD’s ability to handle occlusions, leading to more reliable tissue tracking in challenging surgical environments.

\begin{figure}[h]
\centering
\includegraphics[width=0.9\textwidth]{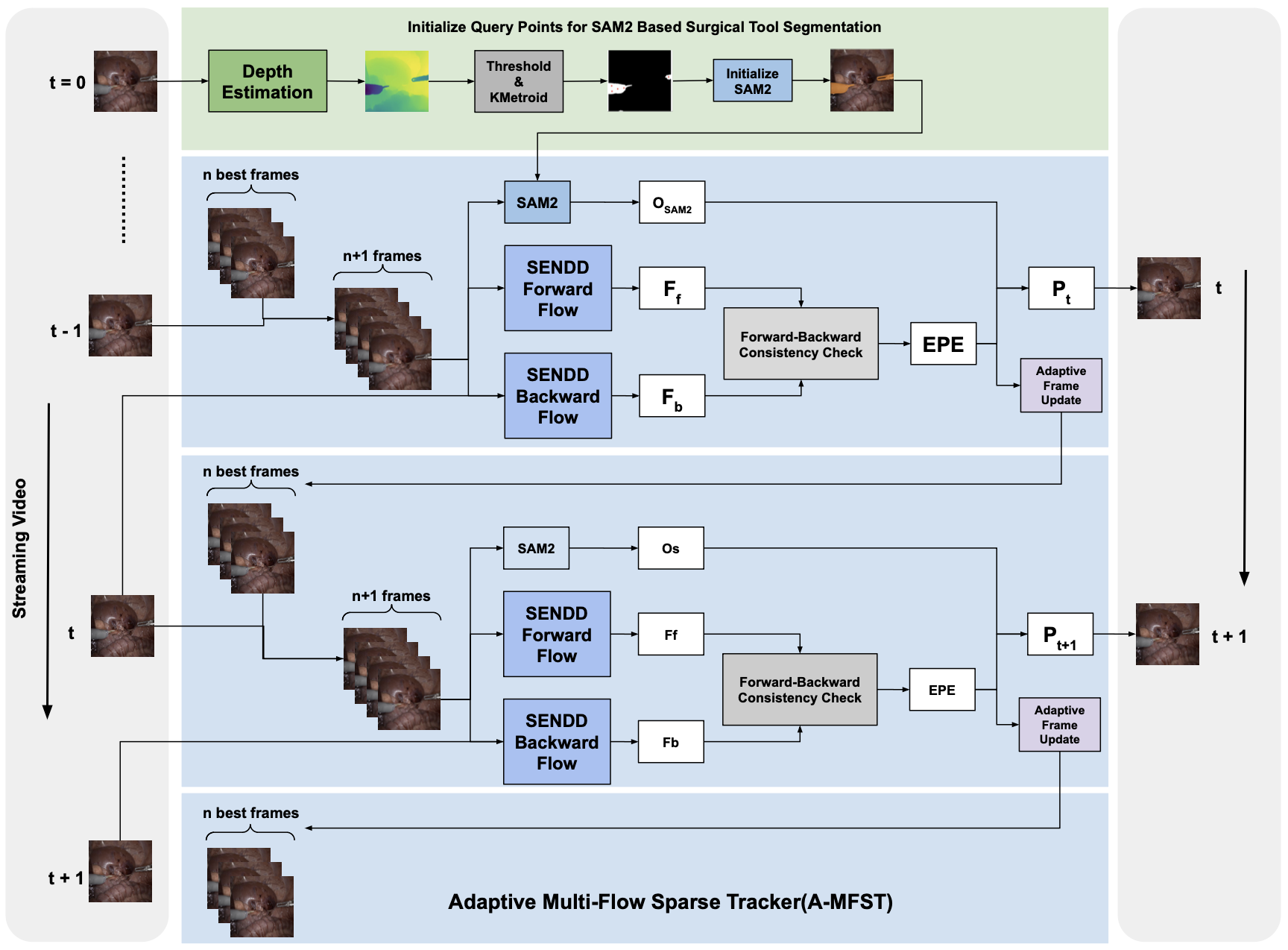}
\caption{Overall structure of the Adaptive Multi-Flow Sparst Tracker(A-MFST)}\label{A-MFST_structure}
\end{figure}

%\subsection{Adaptive Multi-Frame Sparse Track(A-MFST) for Frame Selection and Occlusion Handling}\label{subsec3-1}
\textbf{Adaptive Multi-Frame Sparse Track(A-MFST):} To further improve the performance of MFST, which utilizes fixed intervals for frame selection (e.g., 1, 2, 4, 8, 16, 32, $\infty$), we propose A-MFST~ (Fig.~\ref{A-MFST_structure}). 
 %which dynamically selects the $n$ most reliable previous log-spaced frames based on forward-backward flow consistency. By minimizing the endpoint error (EPE) across candidate frames. A-MFST offers greater flexibility and can increase the speed of tracking by using fewer frames than MFST.
 %to maintain robustness in tracking, particularly in cases of occlusion and non-linear tissue deformations.
%
%\subsubsection{Adaptive Frame Selection and Update via Endpoint Error Minimization}
%
A-MFST dynamically selects the $n_{f}$ most reliable previous frames by selecting the combination of frames that minimizes the sum of endpoint errors from forward-backward flow consistency. 
For each tracking point, we compute the forward-backward EPE from all $n_{f}$ most reliable previous frames and $frame_{t-1}$ to current $frame_{t}$. We then select the combination of $n_{f}$ frames out of these $n_{f} + 1$ frames that minimize the total endpoint error. Each point selects one frame that can minimize its own endpoint error. This strategy, described below in detail, allows for the selection of the most reliable flow estimates for each point, reducing the influence of erroneous or inconsistent flows caused by occlusions.
% In this part, we described the process of assigning tracking points to the optimal frame combinations. This minimizes the total endpoint error, while ensuring occluded points are handled effectively. The A-MFST method combines endpoint error minimization with occlusion detection.

\textbf{SENDD with SAM2 and A-MSFT:}
The endpoint error matrix can be represented by $E \in \mathbb{R}^{(n_{f} + 1) \times n_{p}}$, where $n_{f}$ is the number of most reliable frames 
%and $n_{f}$ best frames + frame[t-1] for optimization, 
and $n_{p}$ is the number of tracking points. Let $O \in \{0,1\}^{(n_{f} + 1) \times n_{p}}$ be the occlusion matrix, where 1 indicates occlusion for a point in a frame. 
$O_{f:p}$ captures whether the predicted position of point $p$ from frame $f$ on current frame is occluded or not. Let $O_{SAM2} \in \{0,1\}^{(n_{f} + 1) \times n_{p}}$ be the occlusion matrix from SAM2, where 1 indicates occlusion for a point in a frame. $O_{SAM2_{f:p}}$ captures whether the predicted position of point $p$ from frame $f$ in the current frame is occluded or not based on the SAM2-segmented tool mask.
An {\em occlusion condition} occurs in the current frame, when the prediction from all the back-checked frames are all in the occluded SAM2 mask or the minimum Endpoint Error is still larger than a threshold as follows:
\begin{equation}
O_{:,p} = 1 \quad \text{if} \quad \forall f \in \{f_{1},\dots, f_{n_{f}}, f_{t-1}\}, \ O_{SAM2_{f,p}} = 1 \ \text{or} \ \min_f E_{f,p} > \tau
\end{equation}
where $\tau$ is a predefined endpoint error threshold, $f$ is the index for each frame, and $p$ is the index for each point. 
If the prediction from all $n_f + 1 $ frames report occlusion or the minimum endpoint error across frames exceeds the threshold $\tau$, the point is marked as occluded in the current frame.

Once a point is marked as occluded, the endpoint error for that point across all frames is set to zero, since this occluded point should not affect the best frame selection for other points. 
The best frame corresponding to this point before occlusion will not be updated and will be remain in the set of $n_f$ most reliable frames to maintain the features of this point, so $E_{:,p} = 0$ if $p$ is occluded.
%
%\begin{equation}
%E_{:,p} = 0 \quad \text{if} \ p \in \text{occluded points}
%\end{equation}

Let $F = \{f_{1}, f_{2},\dots, f_{n_{f}}, f_{t-1}\}$ be the set of $n_{f} + 1$ available frames. The possible frame combinations $C \subset F$ are defined as:
\begin{equation}
C_N = \{ C \subseteq F \ | \ |C| = N \}
\end{equation}

For each combination $C \in C_N$, the endpoint error for the selected frames is computed for each point. The minimum error for each point across the selected frames and the total endpoint error are:
\begin{equation}
e_C(p) = \min_{f \in C} E_{f,p} \,\,\,\, ; \,\,\,\,  E(C) = \sum_{p=1}^{P} e_C(p)
\end{equation}

%The total endpoint error for a combination $C$ is then defined as the sum of the minimum endpoint errors for all points:
%\begin{equation}
%E(C) = \sum_{p=1}^{P} e_C(p)
%\end{equation}

To identify the optimal combination of frames $C^{*}$, we minimize the total endpoint error over all combinations; this ensures that we select the frame combination that minimizes the cumulative endpoint error across all points. Once the optimal combination $C^{*}$ is identified, each point $p$ is assigned to the frame within $C^{*}$ that provides the minimum endpoint error, leading to the frame assignment $f^{*}(p)$:
\begin{equation}
C^{*} = \arg \min_{C \in C_N} E(C) \,\,\,\,\, ; \,\,\,\, f^{*}(p) = \arg \min_{f \in C^{*}} E_{f,p}
\end{equation}

%This ensures that we select the frame combination that minimizes the cumulative endpoint error across all points. Once the optimal combination $C^*$ is identified, each point $p$ is assigned to the frame within $C^*$ that provides the minimum endpoint error. The frame assignment for each point is given by:
%\begin{equation}
%f^*(p) = \arg \min_{f \in C^*} E_{f,p}
%\end{equation}
The adaptive frame update strategy removes occluded points from optimization and selects the optimal combination of frames by minimizing the total endpoint error. By ensuring that occlusion detection is incorporated into the error minimization process, we achieve robust tracking even in the presence of occlusions.

\section{Experimental Results and Discussion}\label{sec1}
This section presents the evaluation of the proposed SAM2-based instrument segmentation, MFST and A-MFST on the STIR dataset~\cite{schmidt2023stir}, which was previously used to evaluate the original SENDD model. The STIR dataset consists of stereo endoscopic videos captured during surgical procedures, annotated with ground truth tracking points for evaluation. 

\subsection{Ablation study of A-MFST}
For the ablation study, we evaluated the contribution of each component in the proposed algorithm when integrated with SENDD, compared to the original SENDD, as shown in Table~\ref{ablation_study_table}. The evaluation metrics include Mean Endpoint Error (MEE), Mean Chamfer Distance (MCD), and \( \delta_{\text{avg}}^{x} \) from TAP-Vid~\cite{doersch2022tap} averaged over accuracy thresholds of [4, 8, 16, 32, 64] pixels as metrics. For evaluation of SAM2, we also calculated MME, \( \delta_{\text{avg}}^{x} \) and the percentage of occluded points below 64 pixels error \( <\delta^{64} \) detected by SAM2. The \( <\delta^{64} \) metric assesses how many occluded points are successfully tracked rather than lost. For those experiments where SAM2 is not part of the main algorithm, SAM2 runs in parallel purely for occlusion evaluation purposes. In the table, the numbers following MFST and A-MFST denote the number of past frames considered for flow prediction in the current frame. For example, MFST7 evaluates frames at indices [0, t-1, t-2, t-4, t-8, t-16, t-32], while A-MFST7 selects the best six frames, along with frame t-1, resulting in seven frames used for flow prediction. We also measured the inference latency (IL) of each method on a desktop equipped with an RTX 4090 GPU.
\begin{table}[th]
\caption{Ablation study of each component in the proposed method with the following metrics: Mean Chamfer Distance(MCD), Mean Endpoint Error(MEE), IL(Inference Latency), \( <\delta_{\text{avg}}^{x} \) and \( <\delta^{64} \) from TAP-Vid~\cite{doersch2022tap}}\label{ablation_study_table}
\begin{tabular*}{\textwidth}{@{\extracolsep\fill}p{2cm}p{1cm}p{1cm}p{1cm}p{1cm}p{1cm}p{1cm}p{1cm}}
\toprule%
& \multicolumn{4}{@{}c@{}}{All Tracking Points} & \multicolumn{3}{@{}c@{}}{Occluded Tracking Points\footnotemark[2]} \\\cmidrule{2-5}\cmidrule{6-8}%
Method & MCD(px) & MEE(px) & \( <\delta_{\text{avg}}^{x} \) & IL(ms) & \( <\delta_{\text{avg}}^{x} \) & \( <\delta^{64} \) & MEE(px) \\
\midrule
SENDD\cite{schmidt2023sendd}  & 45.18   & 22.80  & 66.5 & 50.0  & 22.4 & 56.2 & 78.41\\
SENDD+SAM2                    & 41.99   & 21.25  & 67.8 & 51.6  & 29.3 & 66.6 & 62.87\\
MFST4\footnotemark[1]         & 50.55   & 25.51  & 66.8 & 91.5  & 43.8 & 80.2 & 50.89\\
MFST7\footnotemark[1]         & 38.64   & 19.55  & 68.8 & 135.5 & 47.1 & 83.2 & 46.72\\
A-MFST3\footnotemark[1]       & 44.92   & 22.65  & 67.6 & 68.3  & 44.4 & 77.2 & 49.64\\
A-MFST4\footnotemark[1]       & 40.31   & 20.41  & 69.8 & 79.4  & 45.1 & 77.5 & 49.33\\
A-MFST5\footnotemark[1]       & 39.92   & 20.17  & 70.5 & 92.0  & 45.7 & 74.7 & 49.17\\
A-MFST6\footnotemark[1]       & 38.48   & 19.44  & 71.1 & 106.5 & 46.2 & 80.5 & 46.28\\
A-MFST7\footnotemark[1]       & 38.27   & 19.39  & 71.6 & 120.0 & 48.5 & 85.3 & 43.17\\
\botrule
\end{tabular*}
\footnotetext[1]{Without SAM2.}
\footnotetext[2]{Occluded tracking points detected by SAM2-segmented instrument mask. }
\end{table}

% From the results in Table \ref{tab:method_comparison}, the A-MFST method significantly improved the performance of SENDD across all evaluated metrics. Specifically, the mean EPE for A-MFT is reduced by approximately 34.4\%, demonstrating enhanced accuracy in point tracking. Moreover, tracking robustness improves from 78.4\% for SENDD to 89.5\% for A-MFT. 

\subsection{Comparison with state-of-the-art methods}
\begin{figure}[h]
\centering
\includegraphics[width=0.6\textwidth]{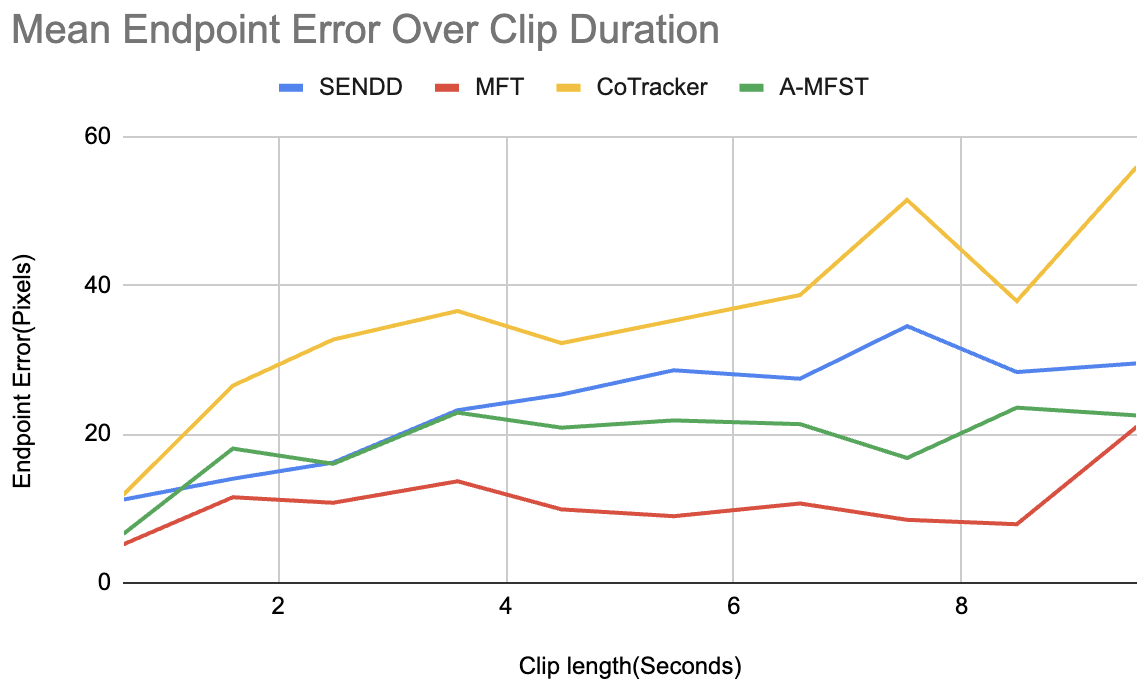}
\caption{Mean Endpoint Error Over Clip Duration.}\label{MEE_over_ClipDuration}
\end{figure}

\begin{figure}[h]
\centering
\includegraphics[width=0.75\textwidth]{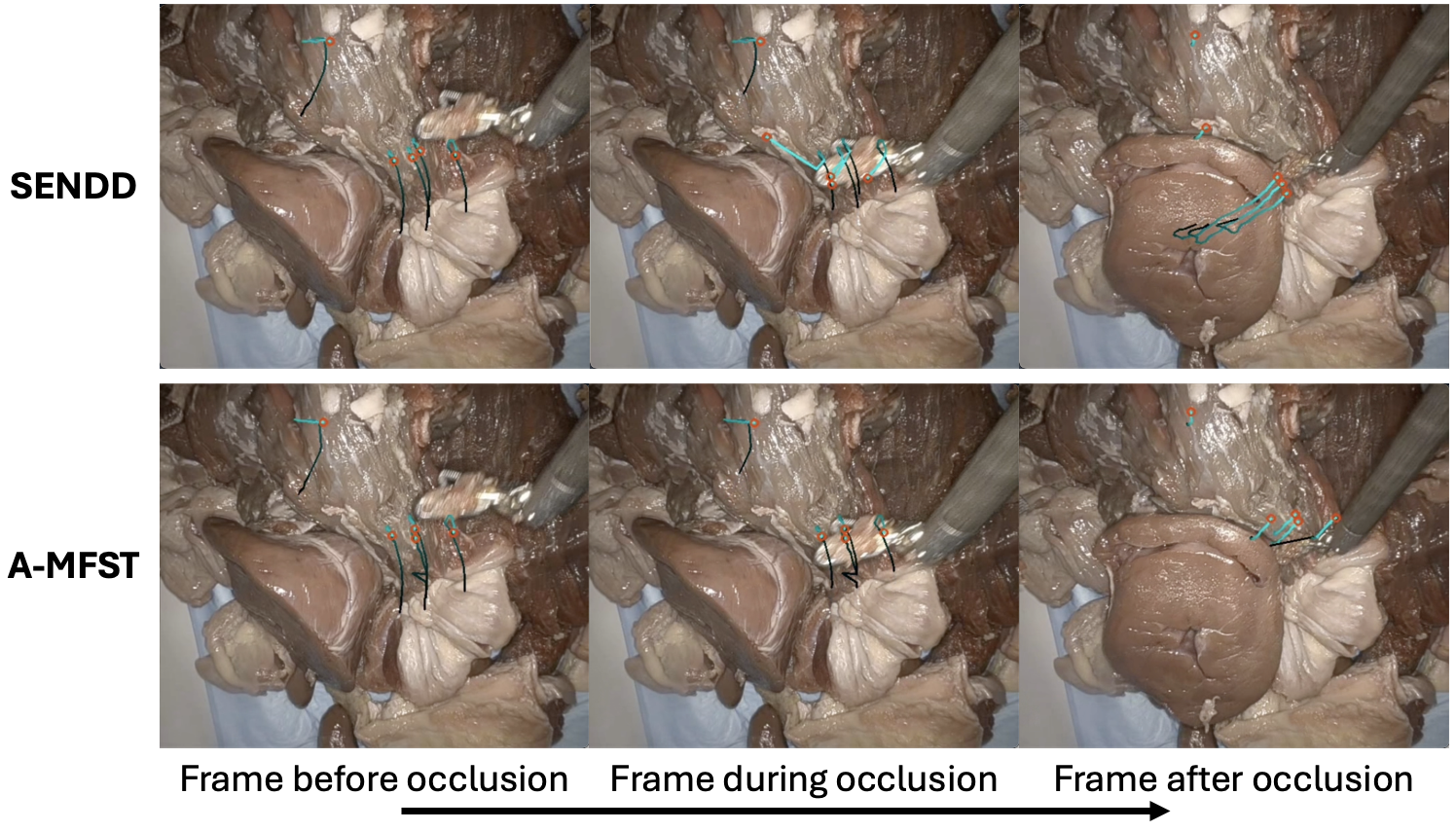}
\caption{Visualization of SENDD and SENDD+A-MFST tissue tracking under occlusion. Red circles are the tracking points on tissue. Blue lines with black tails show the tracking trajectories.}\label{occlusion_demo}
\end{figure}

To evaluate the proposed method with other state-of-the-art methods. We compare A-MFST with SENDD, MFT and CoTracker in Table~\ref{comparison_table}. We use MCD, MEE, \( <\delta_{\text{avg}}^{x} \) as metrics. We also plot the MEE of each method over clip duration in Figure~\ref{MEE_over_ClipDuration} to compare the performance of each method on longer videos. A quantitative visualization example of how SENDD and A-MFST track the points on tissue under occlusion is shown in Figure~\ref{occlusion_demo}.

\begin{table}[h]
\caption{Comparison of A-MFST to other state-of-the-art methods}\label{comparison_table}
\begin{tabular}{@{}lcccc@{}}
\toprule
\textbf{Method} & \textbf{MCD(pixels)} & \textbf{MEE(pixels)} & \textbf{\( <\delta_{\text{avg}}^{x} \)} & \textbf{Inference Latency(ms)}\\ 
\midrule
SENDD\cite{schmidt2023sendd}        & 45.18   & 22.80   & 66.5  & 50.0\\
MFT\cite{neoral2024mft}             & 21.38   & 10.91   & 76.4  & 216.1\\
CoTracker\cite{karaev2023cotracker} & 67.20   & 34.66   & 61.1  & 36.0\\
A-MFST4                             & 39.54   & 20.02   & 70.4  & 80.8\\
\botrule
\end{tabular}
\end{table}

\subsection{Discussion}

The experimental results highlight the effectiveness of the proposed A-MFST in addressing the challenges associated with tissue tracking in surgical environments. By leveraging adaptive multi-frame selection with a forward-backward consistency check and SAM2-based instrument segmentation, the proposed algorithm enhances tracking performance under occlusion, while still running in real-time. %particularly in scenarios where instruments obscure the view of the tissue, while maintaining the ability to apply in real-time.

In Table~\ref{ablation_study_table}, integrating SAM2-based tool segmentation results in a 6\% improvement in MEE tracking accuracy compared to SENDD, with a notable 20\% improvement in MEE and 18\% in \( <\delta^{64} \) for occluded points detected by SAM2. \( <\delta^{64} \) calculated on occluded points reflects how many occluded points are successfully retained during tracking. These improvements highlight the contribution of SAM2 to enhanced occlusion detection, allowing the tracking algorithm to avoid erroneous updates for points temporarily hidden by surgical instruments. However, while occlusion detection is significantly improved, additional advancements like MFST and A-MFST are necessary to effectively recover points once they re-emerge from occlusion.

% By using SAM2 with depth-based initialization, along with automatic occlusion detection, we significantly improve the robustness of tissue tracking in complex surgical scenes where instruments frequently occlude tissue regions. This method minimizes drift and reduces tracking errors, particularly during long occlusions.

In Table~\ref{ablation_study_table}, both MFST and A-MFST demonstrate significant improvements in tracking accuracy. Leveraging a forward-backward consistency check, MFST and A-MFST select the optimal flow path from back-checked frames, ensuring more reliable tracking. The adaptive frame selection mechanism introduced in A-MFST further enhances both tracking performance and processing speed compared to MFST. By selecting the $n$ most consistent frames based on endpoint error (EPE), A-MFST reduces memory usage. In contrast, MFST requires storing information from each frame until the largest logarithmically spaced interval frames have passed, whereas A-MFST only retains information for the $n$ most consistent frames. This approach not only accelerates tracking but also minimizes drift by selecting the most reliable frames for each point, thereby enhancing resilience to occlusions and improving overall tissue tracking accuracy. A-MFST's adaptive design is particularly well-suited for the challenging environment of robotic-assisted surgery, where occlusions and non-linear tissue deformations are frequent.

When comparing A-MFST with and without SAM2, the performance improvement is modest. This is largely due to the overlap between the contributions of the forward-backward consistency check and SAM2 for detecting instrument occlusions. However, SAM2-based instrument segmentation continues to enhance the robustness of occlusion detection in A-MFST.

In Table~\ref{comparison_table}, A-MFST achieves the best performance among state-of-the-art algorithms capable of real-time application. While MFT exhibits the highest tracking accuracy, its inference latency renders it unsuitable for real-time scenarios. A-MFST, on the other hand, offers a flexible trade-off between speed and accuracy by adjusting the number of selected frames. We choose to show the A-MFST4 to compare with other state-of-the-art methods, as it can maintain a high computing speed without sacrificing too much accuracy. A-MFST4 improved 12\% in MEE and 6\% in \( <\delta_{\text{avg}}^{x} \). Additionally, as shown in Figure~\ref{MEE_over_ClipDuration}, A-MFST outperforms other real-time algorithms on longer video sequences and has the closest performance to MFT.

Overall, these findings suggest that the proposed algorithm is a promising approach for real-time tissue tracking in robotic-assisted surgeries, providing a reliable solution for handling occlusions and ensuring high tracking accuracy.

\section{Conclusion}\label{sec1}

In this paper, we presented the A-MFST algorithm, designed to improve tissue tracking in robotic-assisted surgery, particularly under conditions of occlusions and dynamic tissue deformations. By integrating SAM2 for robust instrument segmentation and employing a dynamic frame selection strategy based on forward-backward consistency, A-MFST significantly enhances both tracking accuracy and reliability.

Our experimental evaluation on the STIR dataset demonstrated that A-MFST, in conjunction with SAM2, outperforms the original SENDD method across multiple key performance metrics.  The ablation studies further highlighted the critical contributions of SAM2 segmentation and A-MFST structure in achieving these improvements.

The proposed method not only enhances tissue tracking accuracy but also maintains its suitability for real-time application in surgical environments, where timely and precise feedback is crucial. Future work will focus on further refining the adaptive mechanisms to improve robustness and computational efficiency.

In conclusion, the proposed A-MFST algorithm represents a significant advancement in tissue tracking, offering the potential to enhance both the safety and effectiveness of robotic-assisted surgical procedures.

\section*{Declarations}
\bmhead{Funding} This work was supported by a scholarship held by Y. Chen, and by the C.A. LAszlo Chair held by Professor Salcudean.

\bmhead{Conflict of interest} One of the authors, Adam Schmidt, is affiliated with Intuitive Surgical and received support from the company during the development of SENDD, a fundamental algorithm used in this paper.

\bmhead{Code availability} SENDD is not publicly available. As a result, the code for this paper cannot be made publicly accessible. However, the SENDD methods are described in detail in published articles and can be replicated.

%%===========================================================================================%%
%% If you are submitting to one of the Nature Portfolio journals, using the eJP submission   %%
%% system, please include the references within the manuscript file itself. You may do this  %%
%% by copying the reference list from your .bbl file, paste it into the main manuscript .tex %%
%% file, and delete the associated \verb+\bibliography+ commands.                            %%
%%===========================================================================================%%
\bibliography{sn-bibliography}% common bib file
%% if required, the content of .bbl file can be included here once bbl is generated
%%\input sn-article.bbl

\end{document}